\documentclass[lettersize,journal]{IEEEtran}
\usepackage{amsmath,amsfonts}
\usepackage{graphicx}
\usepackage{algorithmic}
\usepackage{array}
\usepackage[table,xcdraw]{xcolor}
\usepackage{textcomp}
\usepackage{stfloats}
\usepackage{subcaption}
\usepackage{enumitem}
\usepackage{url}
\usepackage[hidelinks]{hyperref}
\usepackage{verbatim}

\def\BibTeX{{\rm B\kern-.05em{\sc i\kern-.025em b}\kern-.08em
    T\kern-.1667em\lower.7ex\hbox{E}\kern-.125emX}}
\usepackage{balance}

\begin{document}
%
\title{Mobile Augmented Reality with Federated Learning in the Metaverse}
\author{~\\
Xinyu Zhou,
Jun Zhao
\thanks{The authors are all with Nanyang Technological University, Singapore. Emails: xinyu003@e.ntu.edu.sg, JunZHAO@ntu.edu.sg}
}

\maketitle

\markboth{}%
{}
%



\IEEEtitleabstractindextext{%
\begin{abstract}
The Metaverse is deemed the next evolution of the Internet and has received much attention recently. Metaverse applications via mobile augmented reality (MAR) require rapid and accurate object detection to mix digital data with the real world. As mobile devices evolve, their computational capabilities are increasing, and thus their computational resources can be leveraged to train machine learning models. In light of the increasing concerns of user privacy and data security, federated learning (FL) has become a promising distributed learning framework for privacy-preserving analytics. In this article, FL and MAR are brought together in the Metaverse. We discuss the necessity and rationality of the combination of FL and MAR. The prospective technologies that support FL and MAR in the Metaverse are also discussed. In addition, existing challenges that prevent the fulfillment of FL and MAR in the Metaverse and several application scenarios are presented. Finally, \textcolor{black}{three} case studies of Metaverse FL-MAR systems are demonstrated.
\end{abstract}

}

\IEEEdisplaynontitleabstractindextext

%
\IEEEpeerreviewmaketitle

\section{Introduction}
\begin{figure*}[t!]
\captionsetup{labelfont={color=black}}
    \centering
    \includegraphics[width=0.78\linewidth]{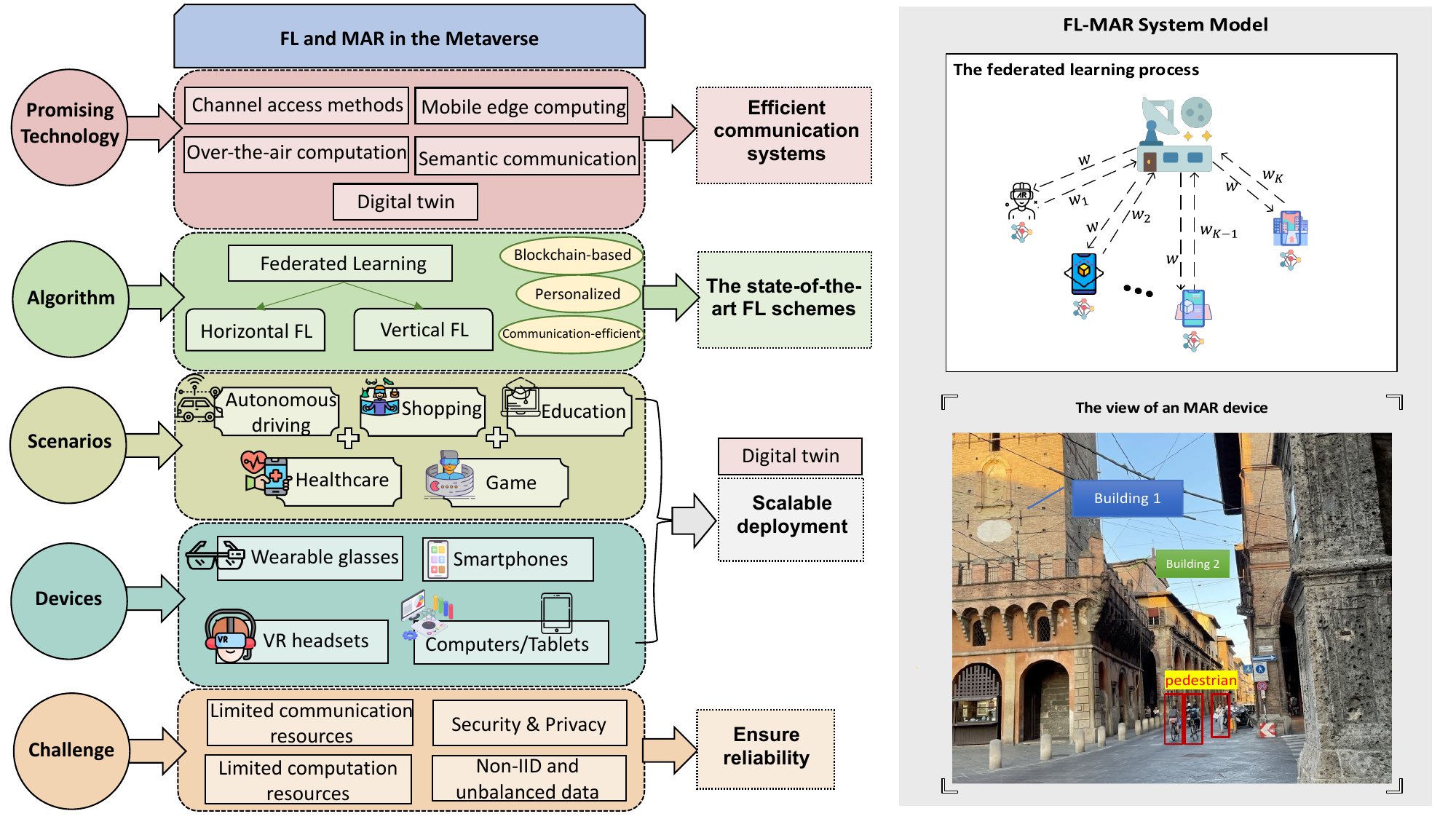}
    \caption{\color{black}The general system model of mobile augmented reality (MAR) with federated learning (FL) in the Metaverse.}
     \label{fig:fl-mar}
\end{figure*}

The Metaverse has become a hot topic recently. 
Mark Zuckerberg made the term famous in 2021 when he announced that Facebook would change its name to Meta and shift its future to build Metaverse technologies. The
Metaverse integrates augmented reality (AR), virtual reality (VR) and 3D technologies to create a fully immersive virtual world.
Mobile augmented reality (MAR) brings the Metaverse to mobile user equipments. 
With the development of mobile technologies, it has been increasingly common for mobile users to leverage AR services to interact and entertain themselves in the real world.
As machine learning technologies are applied to mobile devices, people are developing more intelligent MAR applications in the Metaverse scenarios such as daily communications, entertainment, medical care, travel, transportation, etc.
Fig. \ref{fig:fl-mar} depicts a scenario where users can see descriptions of each building.
Such MAR applications require rapid and accurate object detection to mix digital data with the real world. With the fast development of wearable and mobile devices (e.g., Google Glass, Microsoft Hololens), such a scenario is not a figment of the imagination. Researchers have implemented effective object detection algorithms on mobile devices \cite{cai2021yolobile}.

Usually, current machine learning models require large amounts of data for training to achieve good performance, whereas it is a challenging task to collect those data. As concerns about data security and privacy increase, related regulations and laws are being introduced one after another. Hence, considering the growing difficulty of collecting sensitive datasets to a server for centralized training, distributed learning is a big trend in the future. In an MAR system, devices can train their object detection models separately. However, a mobile device can only store or collect limited data, resulting in a less accurate model. To address this, federated learning (FL) can be incorporated into the MAR system.
FL was proposed by Google in 2017 \cite{mcmahan2017communication}. It allows each device to train a shared model collaboratively without sharing local data with others. As shown in Fig. \ref{fig:fl-mar}, after a few local iterations, each device uploads its local model parameter to a central base station, and the station will send back an updated global model to each device for continuous training. We refer to the system as the FL-MAR system.

Integrating FL into the design of the MAR system poses some challenges. First, limited communication and computing resources can lead to latency between the server and users. Second, achieving a satisfactory model often requires many local iterations and communication times, which also adds a large amount of energy consumption for mobile devices. Moreover, latency and energy consumption are often in conflict. It is necessary to find an appropriate resource allocation strategy to optimize latency and energy consumption. Besides, in the FL-MAR system, the video frame resolution affects the object recognition accuracy, and it influences the computation energy and time when training on each device. 
Hence, to minimize latency and energy consumption and maximize the detection accuracy, we should find how to assign communication and computation resources (i.e., transmission power, bandwidth, CPU frequency and video frame resolution) for each device.


This article first discusses promising technologies for \mbox{FL-MAR} in the Metaverse in Section \ref{sec:tech}. Then, in Section \ref{sec:challen}, we present existing challenges for the applications. Section~\ref{sec:application} demonstrates various application scenarios of FL-MAR in the Metaverse. Finally, Section \ref{sec:case_study} eloaborates three case studies of FL-MAR systems. The \textcolor{black}{system diagram} of FL-MAR is illustrated in Fig.~\ref{fig:fl-mar}.


The contributions are as follows:
\begin{itemize}
\item \textcolor{black}{We are the first to provide the essential design requirements of incorporating FL and MAR into the Metaverse. These design requirements include efficient communication systems, the state-of-the-art FL schemes, scalable deployment and reliability.}
    \item Challenges in the applications of FL-MAR to the Metaverse are also listed from different aspects, including limited communication/communication resources, security and privacy, \textcolor{black}{non-IID and unbalanced data}, etc.
    \item To demonstrate the practicality, we demonstrate case studies of FL-MAR systems in the Metaverse. One is FDMA-enabled, and the other is based on NOMA. \textcolor{black}{We also explore the impact of non-IID, unbalanced data, and image resolutions on model performance.}
\end{itemize}

\section{Enabling Technologies for FL-MAR in the Metaverse} \label{sec:tech}
This section lists some technologies needed for applying FL and MAR to the Metaverse. \textcolor{black}{These technologies have the potential for constructing efficient communication systems. Besides, the state-of-the-art FL algorithms that can contribute to the Metaverse are discussed.} 

\vspace{-10pt}
\subsection{Channel access methods}

\textbf{Frequency Division Multiple Access (FDMA)}. FDMA is a channelization protocol that divides the bandwidth into non-overlapping channels and assign each channel to separate users. 
FDMA is one of the most commonly used analog multiple access methods. It has some advantages: 1) It is technically easy to be implemented. 2) Signals can be transmitted simultaneously while not interfering with each other. 
Moreover, FDMA also has some disadvantages: 1) Bandwidth utilization is limited since channels will be idle if users do not utilize them. 2) If many signals of different frequencies are transmitted simultaneously, \mbox{inter-modulation} distortion is possible to happen at the transponder.

\textbf{Time Division Multiple Access (TDMA)}. TDMA allows different users to share the same frequency by dividing each channel into different time intervals. At each time interval, the frequency is used for one user exclusively. Compared to FDMA, the advantages of TDMA are: 1) The transmission rate is flexible because multiple slots can be allocated to one user. 2) It can handle the changeable bit rate.
However, the disadvantages include the implementation complexity and the requirement of synchronization.

\textbf{Non-orthogonal Multiple Access (NOMA)}. NOMA has been seen as a promising technology for intensifying the throughput in future wireless systems. Unlike conventional orthogonal multiple access (OMA), it enables multiple users on the same channel to be multiplexed to maximize the throughput and lower the latency of the system. It adopts superposition coding at the transmitter and utilizes successive interference cancellation (SIC) at the receiver to distinguish signals of users. Hence, it increases OMA's rate region.

For other channel access schemes, such as CDMA, SDMA and OFDMA, interested readers can refer to \cite{kumar2020multiple}.

 \vspace{-5pt}
\subsection{Semantic Communication}
Semantic communication has been deemed the breakthrough of Shannon's paradigm as it transmits only the relevant semantic information about the specific resource. It does not aim at the accurate transmission of bit sequences, but rather focuses on transmitting information in the form of semantic structures \cite{xu2022full}.
Hence, the data traffic can be significantly lowered, which is why it can be one of the promising solutions leading to an efficient communication network in the Metaverse.
Studies of semantic communication for the Metaverse are in the early stage~\cite{xu2022full}. 
\textcolor{black}{One of the challenges of FL-MAR in the Metaverse is the  high communication overhead, which is the bottleneck of FL itself. Considering the scenario of MAR, there might be more frequent communication. Therefore, to overcome this bottleneck, a potential solution may be to not transmit the raw data during the communication but instead use semantic communications.}

\vspace{-5pt}
\subsection{Over-the-Air Computation }
Over-the-air computation (AirComp) enables computation function by adding the analog wave in a multiple-access channel. By utilizing the interference for implementing the computation function, the wireless channel can be used as a computer. In a distributed system, the signals sent by mobile devices are superposed over the air and aggregated by the receiver as a weighted sum. The weights represent channel coefficients \cite{zhu2021over}. 
In the Metaverse, numerous devices are connected to communicate through the communication network. Large amounts of data are transmitted by various devices simultaneously while devices wait for immediate feedback. The advent of over-the-air computation may help to build low-latency communication networks in the Metaverse.

\vspace{-10pt}
\subsection{Mobile Edge Computing}
Before the emergence of edge computing, cloud computing was the new paradigm of computing. 
Generally, clouds are servers (e.g., data centers) that can be approached through the Internet. However, such servers are usually located far away from user devices, resulting in long latency. In 2014, European Telecommunications Standard Institute (ETSI) proposed the concept of mobile edge computing (MEC). In the IT service environment, MEC equips cloud-computing capabilities at the edge of mobile networks in the vicinity of mobile users. It aims at reducing latency and providing high-efficient network operations and services \cite{xu2022full}. 
MEC is deployed at access points, such as small base stations, edge servers, users' computers, etc. FL-MAR systems consist of massive mobile devices. Hence, if utilizing the idle computing resources of mobile resources through MEC, the energy consumption and latency in communication networks in the Metaverse can be significantly saved.

\vspace{-10pt}
\subsection{Blockchain and Cryptocurrency}
Blockchain, which is a distributed ledger, first appeared in 1991. 
It stores digital data in blocks, and the blocks are strung together via cryptography. The data is time-irreversible by leveraging cryptographic hash functions. With the merits of transparency and security, blockchain has numerous prospective applications in finance, government, commerce, etc. \textcolor{black}{In the Metaverse, blockchain technology can be utilized to build a privacy-preserving FL framework, as shown by Kang \textit{et al.} \cite{kang2022blockchain}. Blockchain can keep the records of FL model updates transparently while providing a secure approach to aggregating models.}
To protect user privacy, cryptocurrencies are also essential in the Metaverse. Cryptocurrencies (e.g., Bitcoin, Litecoin, Ethereum) are powered by blockchain. 
Additionally, cryptocurrencies are not physical, they exist only in the decentralized network, and their creation is determined by an algorithm (or protocol).
\textcolor{black}{Therefore, cryptocurrencies or some other tokens can be utilized by blockchain as rewards to encourage users to participate in FL. Additionally, blockchain can store the raw data and the digital assets generated by MAR users in a secure and transparent environment, leading to a decentralized virtual management world.}

{\color{black}
\vspace{-13pt}
\subsection{Digital Twin}
Digital twins represent the virtual models of the physical objects within the Metaverse by leveraging advanced modeling technologies. They monitor and simulate physical objects in the real world with real-time data.
Besides, the data from physical objects can be utilized and adapted for self-learning in the Metaverse \cite{wang2023survey_meta}.
Digital twins have already been used in numerous scenarios, including manufacturing, healthcare, etc. Moreover, digital twins contribute to the real-time prediction and maintenance of a system. For example, complex engine systems can be simulated by digital twins to figure out maintenance cycles. Hence, digital twins can help integrate physical assets and the virtual environment, leading to the permeation of the Metaverse in each aspect of life.

\vspace{-10pt}
\subsection{The State-of-the-Art FL Schemes}

FL schemes can be classified from different perspectives, such as data distribution,  privacy \& security, communication strategies, aggregation strategies, application scenarios, etc. This article mainly focuses on two perspectives: data distribution and privacy \& security. 
    
     \textit{Data distribution}. The FL schemes can be categorized by the data distribution into horizontal FL, vertical FL and federated transfer learning. Horizontal FL refers to scenarios that the features of distributed data on each client are similar, but the data are not the same. The FL proposed by Google \cite{mcmahan2017communication} belongs to this category.
     Compared to horizontal FL, the data of vertical FL does not share the same features. 
    Federated transfer learning can be applied to scenarios with insufficient users and overlapping features \cite{FLMetaverse2023}.
    
    \textit{Privacy \& security}. The future Metaverse world involves massive data interactions, and user privacy and data security issues will be critical. Some techniques, such as differential privacy, secure multi-party computation and consensus mechanism, can be utilized with FL to enhance privacy:\\
1). \textbf{Differential privacy (DP)}. It can be seen as a way of the mathematical definition of privacy, which adds noise to personally sensitive information.
    DP ensures that the individual information in the database will not be compromised. The combination of DP and FL is by injecting noise to participating devices' uploaded parameters \cite{MOTHUKURI2021619}.\\
    2). \textbf{Secure multi-party computation (SMC)}. It is a cryptographic protocol that several parties jointly compute an agreed function without exposing each party's data.
    In SMC, data can be shared distributedly with other parties without needing a third-party organization and is still protected. 
    By utilizing SMC in FL, the uploaded parameters from clients are encrypted, but it is expensive if used in a large distributed environment \cite{MOTHUKURI2021619}. SMC is not a single protocol but a collection of technologies, such as Homomorphic Encryption.\\
 3). \textbf{Consensus mechanism}. 
    It could be any mechanism that is \mbox{fault-tolerant} and usually used in blockchain systems to reach a consensus on a data value or a network state among distributed systems (e.g., cryptocurrencies).
    In FL, a consensus mechanism can help coordinate, verify, and achieve an agreement on the updates from various clients \cite{kang2022blockchain}. For example, in the blockchain-based FL, blockchain can keep records of model updates and build a secure and transparent environment.

\textit{Other asepcts}. 
The aforementioned FL schemes do not conflict with each other and can often be combined to meet specific requirements.
Besides, in the Metaverse and MAR applications, personalized FL can be devised to provide personalized contents and services for each user \cite{FLMetaverse2023}. Usually, communication efficiency will be stressed in FL due to the limited communication resources, and thus resource allocation is integrated into the FL system \cite{yang2020energy,zhou2022resource}.
}

\section{Challenges for FL-MAR in the Metaverse} \label{sec:challen}
This section discusses the potential challenges of FL-MAR systems in the Metaverse.

\vspace{-10pt}
\subsection{Limited communication resources}
The demands of bandwidth in the Metaverse are much more significant than current ordinary games and social entertainment, because the Metaverse virtual world has to render massive surroundings (e.g., trees, flowers), buildings, avatars (or people), etc., simultaneously. 
Furthermore, to achieve the immersive experience in the Metaverse, haptic technology is indispensable to create the experience of touch and receive the reactions of the virtual world. However, current communication resources cannot fulfill the need for immediate haptic feedback. Thus, limited communication resources are one of the barriers to the widespread deployment of the Metaverse. To address the limitation, technologies mentioned in Section \ref{sec:tech} could be utilized, such as semantic communications and MEC. 

\vspace{-10pt}
\subsection{Limited computation resources}
In the Metaverse, the resource-constrained devices worn by users not only have to train their own FL model but also are responsible for processing newly generated data and converting the raw data into the 3D virtual world. Apparently, today's mobile devices do not have the computing capability to finish those complex tasks efficiently.
This existing obstacle might be addressed by better MEC mechanisms. MEC assists applications that have requirements of low latency and high bandwidth close to the data source.
Since edge servers are much closer to users than cloud servers, they could provide services with low latency, which is suitable for the Metaverse to provide a real-time and stable immersive VR/AR experience. 
However, if the computing tasks are complex and energy-consuming, they can be uploaded to cloud servers, because
they provide robust computing and storage resources. The appropriate combination of edge- and cloud-based applications is essential to maximize the system performance.

\vspace{-10pt}
\subsection{Security and Privacy}
Our physical world is being transformed into a digital one as time passes. In the Metaverse, people's lives are changing in the areas of shopping, education, tourism, medical care, etc.
There will be new forms of security risks, including the threats of
scams, identity leakage, and data protection.
Although FL can protect user privacy and data security to a certain extent, users in the Metaverse still expose their sensitive information to the virtual world. Besides, in FL, the model parameter transmission has the potential to leak user privacy. 
{\color{black} Besides, future VR/AR in the Metaverse also has the potential to leak user privacy. Table \ref{tab:threats} concludes that the main threats exist in FL and VR/AR in the Metaverse. Albeit virtual avatars in the Metaverse will not reveal the individual's appearance, the actions made by the virtual avatars, the data collected from sensors and threats of various third-party applications can still allow attackers to infer the sensitive information, such as height, age, gender, ethnicity, etc., about individuals.

\begin{table*}[t!]
\captionsetup{labelfont={color=black}}
\centering
\caption{\color{black} Threats in Federated Learning and VR/AR (references \cite{MOTHUKURI2021619,nair2022exploring})}
\begin{tabular}{|
>{\columncolor[HTML]{F8DD9B}}c cc|}
\hline
\multicolumn{3}{|c|}{\cellcolor[HTML]{EBF4FC}Threats in Federated Learning and VR/AR}                                                                                                                    \\ \hline
\multicolumn{1}{|c|}{\cellcolor[HTML]{EBF4FC}}                          & \multicolumn{1}{c|}{\cellcolor[HTML]{C4DEF5}Federated Learning}                                                                                                                                      & \cellcolor[HTML]{C4DEF5}VR/AR                                                                                                                     \\ \hline
\multicolumn{1}{|c|}{\cellcolor[HTML]{F8DD9B}Source of Vulnerabilities} & \multicolumn{1}{c|}{\begin{tabular}[c]{@{}c@{}}Non-secure communication channels\\ Data manipulations\\ Vulnerable aggregation algorithms\end{tabular}}                                & \begin{tabular}[c]{@{}c@{}}Raw data on hardware\\ Client-side applications\\ Third-party servers\\ End users of the same application\end{tabular} \\ \hline
\multicolumn{1}{|c|}{\cellcolor[HTML]{F8DD9B}Privacy Threats}           & \multicolumn{1}{c|}{\begin{tabular}[c]{@{}c@{}}The inference of training data\\ Data leakage/reconstruction from other clients\\ GANs (generative adversarial network)-based inference\end{tabular}} & \multicolumn{1}{l|}{\begin{tabular}[c]{@{}l@{}}Inference of personal information about \\ age, gender, height, ethnicity, etc.\end{tabular}}      \\ \hline
\multicolumn{1}{|c|}{\cellcolor[HTML]{F8DD9B}Security Threats}          & \multicolumn{2}{c|}{\begin{tabular}[c]{@{}c@{}}Data/model poisoning, data tampering,\\ GANs-based attacks, eavesdropping,\\ system malfunction, backdoor attacks\end{tabular}}                                                                                                                                                                           \\ \hline
\end{tabular}\label{tab:threats}
\end{table*}

\vspace{-10pt}
\subsection{Non-IID and Unbalanced Data}
The non-IID and unbalanced data distribution is a long-existed challenge for distributed learning.
The purpose of FL was to mitigate the negative impact of non-IID and unbalanced data distributions \cite{mcmahan2017communication}, but non-IID data still affect the model performance and convergence. Besides, MAR applications involve a variety of data sources, such as raw sensor data, videos, images, etc. Hence, the data heterogeneity results in the difficulty of processing the data for training and updating models.
Effective segmentation methods and datasets may solve different non-IID situations effectively \cite{FLMetaverse2023}.
Moreover, higher image resolution will result in higher model accuracy but with higher energy consumption and training time. Therefore, how to adequately adjust the image resolution for each mobile device is also a challenge. A more detailed case study about the influence of non-IID, unbalanced data and image resolution in FL-MAR systems is provided in Section \ref{subsec:non_iid}.
}


\section{Applications of FL-MAR in the Metaverse} \label{sec:application}

\begin{figure*}[t!]
\captionsetup{labelfont={color=black}}
    \centering
    \includegraphics[scale=0.33]{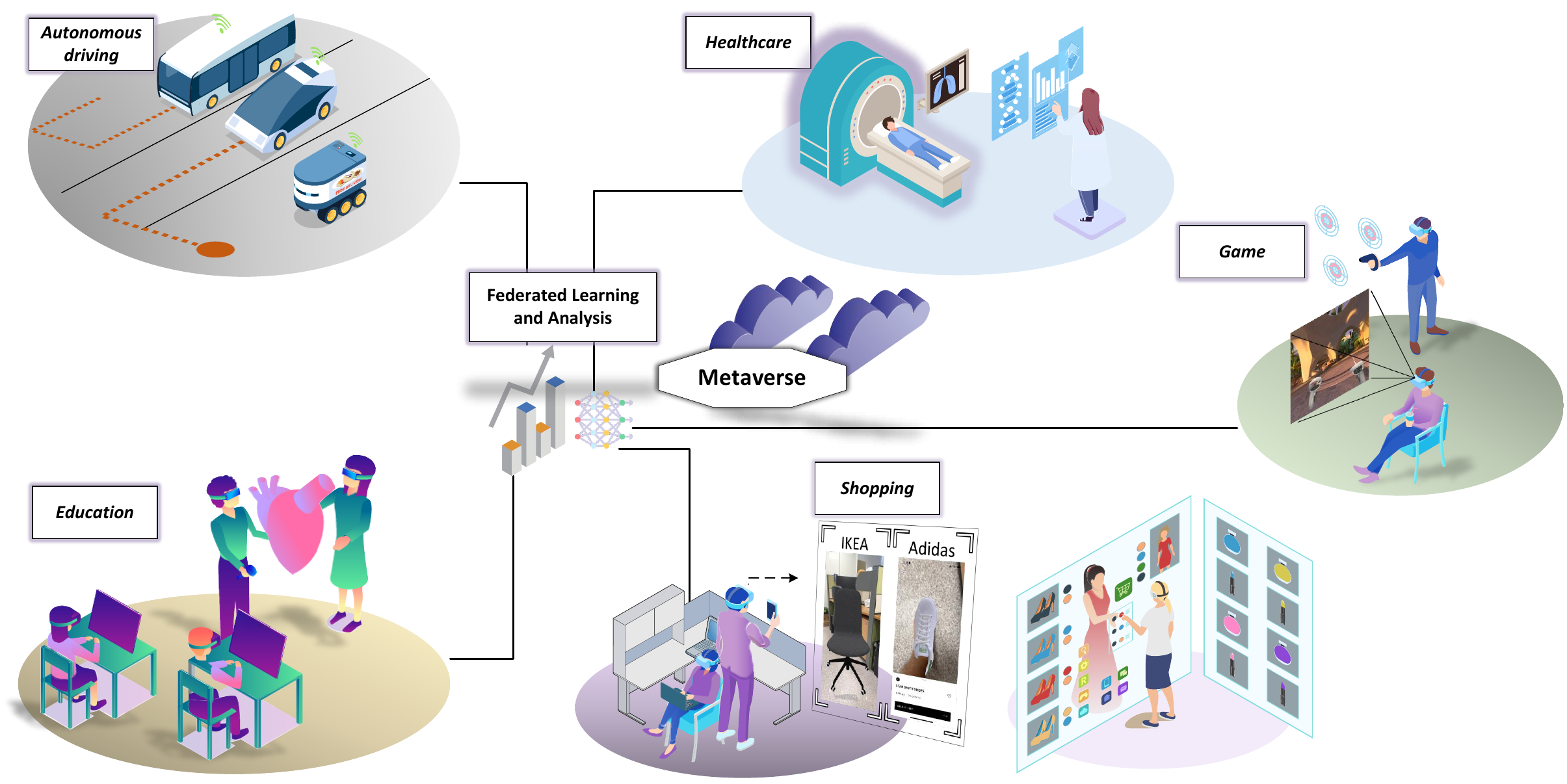}
    \caption{\color{black}The application scenarios of FL and MAR in the Metaverse.\vspace{-10pt}}
    \label{fig:scenarios}
\end{figure*}

This section lists some scenarios in which MAR and FL are applied to the Metaverse. Fig. \ref{fig:scenarios} depicts the scenarios including autonomous driving, shopping, education, healthcare and game playing.

\vspace{-10pt}
\subsection{Autonomous Driving}
Autonomous vehicles are becoming a feasible solution for transportation in the future. Recently, deep learning has been a popular approach to the application of autonomous driving in terms of object detection, obstacle avoidance, etc.
Considering the fact that the capabilities of hardware storage and computation are improving, training models locally is not only beneficial for data security and user privacy but also reduces network energy consumption and latency.
Since different cars experience various environments, such as weather and lighting conditions, incorporating FL into this scenario will help each vehicle build a more accurate model.
Additionally, since Metaverse has immediate physical-virtual world interaction characteristics, it can simulate various kinds of driving situations, including some rare cases. Therefore, it helps
test whether self-driving cars are safe and reliable in various extreme conditions.

\vspace{-10pt}
\subsection{Shopping}
Online shopping has become a part of people's lives.
As the number of MAR applications grows, online shopping also enjoys this convenience. For instance, IKEA, the company that sells ready-to-assemble furniture, appliances and home services, equips its app with AR capabilities that allow users to place 3D models of equal scale with the real size virtually in their homes.
Adidas also launched the AR footwear try-on function in its iOS app. \textcolor{black}{The shopping scenario of \mbox{Fig. \ref{fig:scenarios}} illustrates both two examples of MAR applications}.


However, it is evident that the MAR applications still need improvement. Each person has different looks and living places. Therefore, by incorporating FL, the mobile device can learn its own model to fit the specific person. Additionally, in the Metaverse, people will reveal similarities to their real lives in the virtual
worlds due to more time spent virtually. 
Products such as digital clothing, furniture, cosmetics and so forth may have a similar status to purchases in the real world.

\vspace{-10pt}
\subsection{Education}
The Metaverse will transform the educational environment in the future \cite{zhang2022metaverse}. Different from traditional in-person learning and online learning, Metaverse-based learning will be an environment which is a mixture of the virtual and real world. It allows students to interact with each other in a virtual and decentralized setting and join various complex learning activities. In light of the profusion of online learning experiences during the Covid-19 period, Metaverse-based learning is much needed now. For example, in geography classes, students can immerse themselves through VR headsets in geography and experience the differences between different climates in different regions.

{\color{black}
\vspace{-5pt}
\subsection{Healthcare}
Intelligent healthcare has been studied by scholars in biomedical engineering in recent years. For example, in 2022, a study envisioned the benefits the Metaverse could bring to the healthcare domain \cite{wang2022development_health}. 
With the emergence of virtual clinics and hospitals, doctors could diagnose patients remotely using avatars. For example, after a CT scan, if a patient is diagnosed with heart disease, the patient avatar can simulate several possible pathologies by using different virtual scanners with the help of digital twins \cite{wang2022development_health}. Other than virtual diagnosis, the Metaverse gives another possibility to medical training. The trainee surgeons can practice surgeries on \mbox{3D-constructed} bodies, which helps to decrease the risk of actual surgeries.
Besides, the raw data generated by various patients will raise privacy concerns. FL and DP can be incorporated to enable the development of AI models used in the healthcare domain and lower the risk of data breaches and privacy violations.

\vspace{-5pt}
\subsection{Game}
As the game-playing scenario illustrated in Fig. \ref{fig:scenarios}, users will have an immersive game-playing experience in the Metaverse. Unlike common AR/VR games, the game playing in the Metaverse will stress the importance of immersion, the low latency will provide real-time feedback, and multiplayer interactions will be smoother than today's AR/VR games. Although there are some Metaverse-related games such as Decentraland and the Sandbox on the market, they do not truly achieve an immersive gaming experience. Instead, they utilize the Metaverse worldview to allow users to create contents in the form of Non-Fungible Tokens (NFTs) and make profits by using cryptocurrencies.
Besides, FL can bring personalized gaming experiences depending on each player's preferences, behaviors and interests. These raw data could feed into the training of game AI. For example, the non-player characters can talk with different players in different emotions and tones based on their personalities.

}

\vspace{-2pt}
\section{Case Studies of FL-MAR in the Metaverse} \label{sec:case_study}
In this section, we present three case studies of FL-MAR in the Metaverse. One FL-MAR system's channel access scheme is FDMA, and the other one uses NOMA. 
We also study the impact of non-IID, unbalanced data distributions and image resolution on the model performance of the FL-MAR system. 


\subsection{FDMA-enabled FL-MAR in the Metaverse} \label{sec:fdma}
First, we investigate a basic FL-MAR system via FDMA. In \cite{zhou2022resource}, we formulate a weighted sum of total energy, time consumption and accuracy by using three weight parameters.
We optimize the allocation of the bandwidth, transmission power, CPU frequency setting and MAR video frame resolution for each participating device in the Metaverse. By setting different weight parameters, our resource allocation algorithm can adapt to different requirements of the FL-MAR system, either \mbox{time-sensitive} or energy-hungry.

\begin{figure}[t!]
    \centering    \includegraphics[width=0.49\textwidth]{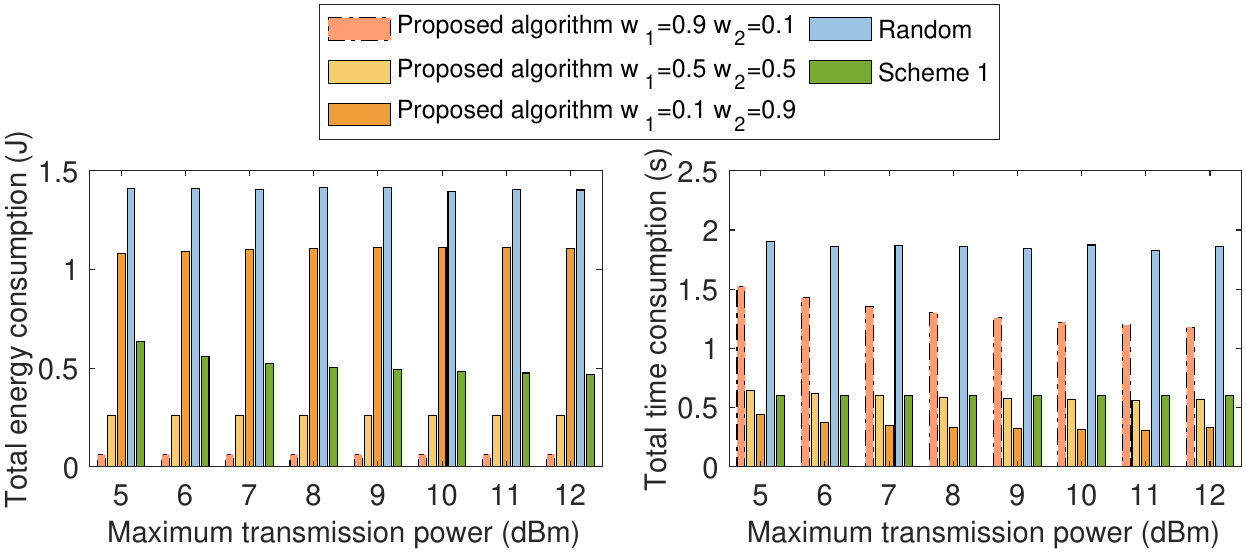}
    \caption{FDMA-enabled FL-MAR system: simulation results under different transmit power limits. The local epochs and global communication round are $10$ and $1$. \vspace{10pt}
    }
    \label{fig:fdma_et_cost}
\end{figure}

We assume there are $40$ users in the system. Fig. \ref{fig:fdma_et_cost} contains two subfigures. One shows the total energy consumption, and the other is the total time consumption. We choose three pairs of weight parameters to compare our resource allocation algorithm with a random allocation strategy and \textcolor{black}{the algorithm proposed by Yang \emph{et a.l} \cite{yang2020energy}} (we name it ``Scheme 1'') under different maximum transmit power limits. Note that $w_1$ is the weight parameter of energy consumption, and $w_2$ is the weight parameter of time consumption. The weight parameter of the model accuracy is fixed because we focus on the energy and time consumption here.  
If $w_1$ (resp., $w_2$) becomes larger, the
proposed algorithm will emphasize minimizing the energy cost (resp., time consumption). Hence, it can be seen obviously from each bar that as $w_1$ (resp. $w_2$) increases, the energy (resp. time) consumption will decrease. 
\textcolor{black}{Besides, although Scheme 1 has roughly the same time consumption as our proposed algorithm ($w_1=w_2=0.5$), our proposed algorithm ($w_1, w_2$)=($0.5, 0.5$) and ($0.9, 0.1$) still achieve a better result in terms of energy consumption compared to Scheme 1.}

\vspace{-10pt}
\subsection{NOMA-enabled FL-MAR in the Metaverse}
\begin{figure}[t!]
    \centering
    \includegraphics[width=0.48\textwidth]{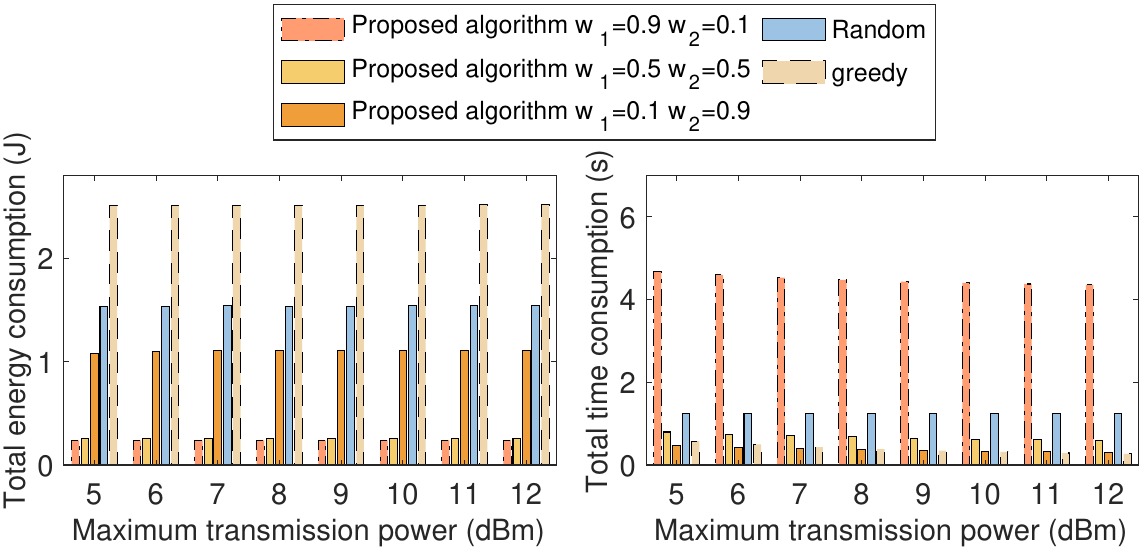}
    \caption{NOMA-enabled FL-MAR system: simulation results under different transmit power limits. \textcolor{black}{The local epochs and global communication round are $10$ and $1$}.\vspace{10pt}}
    \label{fig:noma_et_cost}
\end{figure}

Here, we study the NOMA-enabled FL-MAR system in the Metaverse.
We also devise a resource allocation algorithm and show the validity of our algorithm, which jointly optimizes the weighted sum of energy and time consumption. Assume there are $40$ users and $20$ channels. There are $2$ users multiplexed on one channel.
In Fig. \ref{fig:noma_et_cost}, we compare three pairs of weight parameters $(w_1, w_2)=(0.9,0.1),(0.5,0.5)$ and $(0.1,0.9)$ with a random allocation strategy \textcolor{black}{and a greedy allocation strategy}.
$w_1$ and $w_2$ have the same meaning as that in Section \ref{sec:fdma}.

Moreover, Fig. \ref{fig:noma_et_cost} contains comparisons under different maximum transmission power limits of the total energy consumption and time consumption.
It can be concluded that when the maximum transmission power increases, total time consumption slightly decreases.
Our resource allocation algorithm performs better than the random allocation strategy in the aspect of energy optimization.
In terms of total time consumption, the proposed algorithm performs worse than the random allocation when $(w_1=0.9, w_2 = 0.1)$, because the case of $ w_1=0.9 $ emphasizes more about the energy optimization and less about the time minimization. The simulations show the effectiveness of our approach for different weight parameters. \textcolor{black}{Besides, it could be observed that the greedy allocation strategy focuses most on optimizing time consumption. Thus, in terms of time optimization, it is almost the same as when we have $(w_1=0.1, w_2=0.9)$, but the energy consumption is even worse than the random strategy.}

\vspace{-10pt}
\subsection{Analysis of the difference between FDMA-enabled and NOMA-enabled FL-MAR system}
It could be concluded from Fig. \ref{fig:fdma_et_cost} and Fig. \ref{fig:noma_et_cost} that, in terms of total energy consumption, there is little difference between the performance of FDMA-enabled and NOMA-enabled FL-MAR system. Regarding total time consumption, the FDMA-enabled system performs slightly better when $w_1=0.9$ and $w_2 = 0.1$. When $(w_1=0.5, w_2=0.5)$ and $(w_1=0.1, w_2=0.9)$, their performance is similar. In addition, the random scheme under NOMA outperforms the random scheme under FDMA. From the simulation results and our theoretical analyses, when the system has ample channel resources, careful optimization of FDMA has comparable performance with NOMA, so there is no need to implement the more complex NOMA for resource-rich scenarios. Yet, when the system has limited channel resources, NOMA can improve the system performance by leveraging power-domain orthogonality.
Our findings are also consistent with recent results in~\cite{li2022exploiting}. 
We hope our preliminary simulation can motivate more real-world NOMA experiments for the Metaverse in the research community.

\begin{figure*}[t!]
\captionsetup{labelfont={color=black}}
    \centering
    \begin{subfigure}{0.42\textwidth}
    \centering
\includegraphics[width=0.8\linewidth]{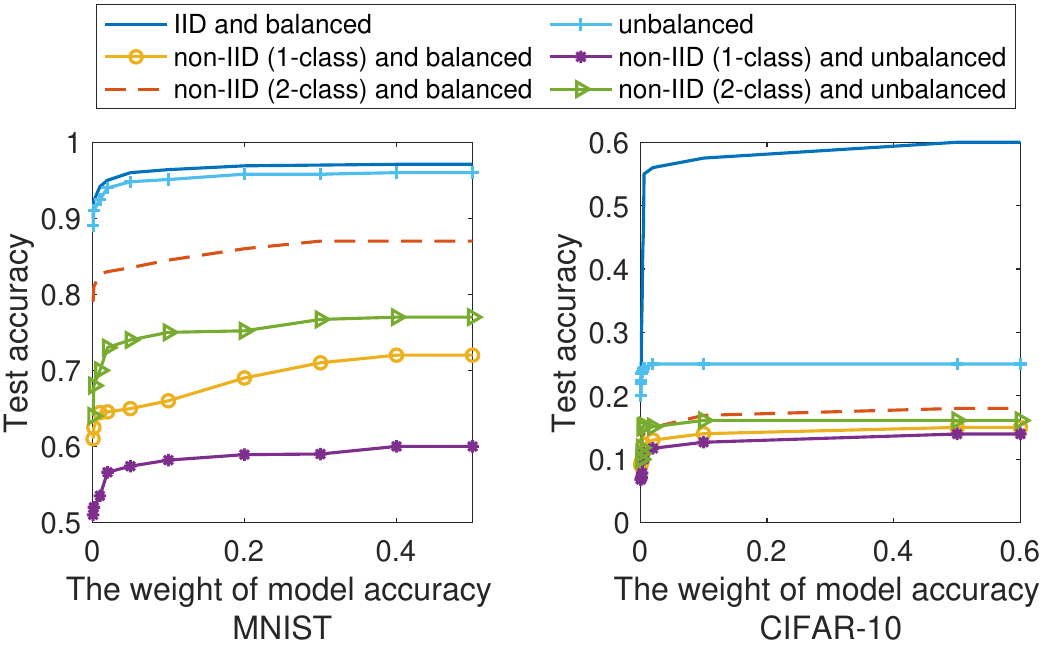}
    \caption{\color{black}The test accuracy under different weights of model accuracy for MNIST and CIFAR-10 datasets. MNIST: The options of the image resolution are $\{7^2, 14^2, 28^2\}$ pixels. CIFAR-10: The options of the resolution are $\{8^2, 16^2, 32^2\}$ pixels. The number of local epochs and global communication rounds are $5$ and $100$. The number of clients is $10$. The models are CNNs (convolutional neural networks).} \label{fig:mnist_cifar}
    \end{subfigure}
        \hspace{5pt}
    \begin{subfigure}{0.53\textwidth}
    \centering    \includegraphics[width=0.52\linewidth]{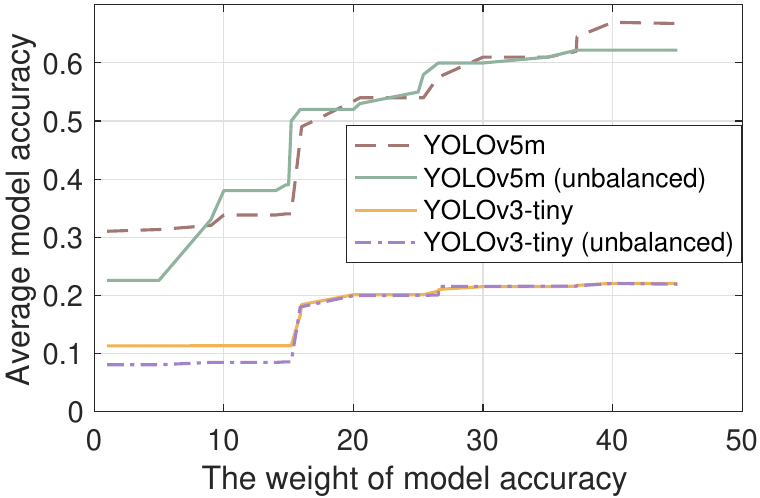}
    \caption{\color{black}The model accuracy under different weights of model accuracy for COCO dataset: The options of the resolution are $\{160^2, 320^2, 480^2, 640^2\}$ pixels. The model accuracy refers to the mean Average Precision, which is a widely used measurement in object detection. The number of local epochs and global communication rounds are $5$ and $50$. The number of clients is $10$. The models are YOLOv5m and YOLOv3-tiny (i.e., the real-time object detection algorithms).}
    \end{subfigure}
    \caption{\color{black}Experiments of image classification and object detection. \vspace{-13pt}
    }
    \label{fig:acc}
\end{figure*}

{\color{black}
\vspace{-10pt}
\subsection{Non-IID and unbalanced data distribution in FL-MAR} \label{subsec:non_iid}
In this section, we study the impact of non-IID, unbalanced data distributions and the image resolution on the model performance. 
In \cite{zhou2022resource}, other than energy and time, we also optimize the model accuracy.
Fig. \ref{fig:acc} depicts the relationship between model accuracy and the weight of model accuracy, and the basic experimental settings are listed in it.  In Fig. \ref{fig:acc}(a), ``non-IID (1-class)'' indicates that every client possesses data with one unique label. ``non-IID (2-class)'' suggests that each client's data includes two different labels. The unbalanced situation is randomly assigning samples to each client.
It could be observed that in both Fig. \ref{fig:acc}(a) and (b), as the weight of model accuracy increases, the model accuracy improves because the higher image resolution used for training will be selected. Additionally, in Fig. \ref{fig:acc}(a), the ``non-IID (1-class) and unbalanced'' has the worst performance in MNIST and CIFAR-10. Compared to the unbalanced setting, the impact of non-IID is more significant. Moreover, the image resolution has a more substantial influence on CIFAR-10's accuracy compared to MNIST, and this may be caused by the higher complexity of CIFAR-10. Fig. \ref{fig:acc}(b) also reveals the fact that the unbalanced data will result in a worse model.  
}

\vspace{-10pt}
\section{Conclusion}
In conclusion, this article gives insights into the necessity and rationality of a federated learning enabled mobile augmented reality system (FL-MAR) in the Metaverse.
The combination of FL and MAR in the Metaverse not only helps protect user privacy to a certain extent but also utilizes the available computing resources on mobile devices. 
Besides, this article lists and explains the promising technologies that enable FL-MAR systems in the Metaverse.
Some application scenarios are also given in this article, including autonomous driving, shopping, education and so forth.
Finally, three case studies are evaluated. The energy and latency are analyzed for FDMA-enabled and NOMA-enabled FL-MAR systems. Also, we study the impact of non-IID, unbalanced data and image resolution on the FL-MAR system. We envision our paper to motivate more research on leveraging FL for the Metaverse, and designing more efficient channel access mechanisms to enable the Metaverse for mobile user equipments.





\ifCLASSOPTIONcaptionsoff
  \newpage
\fi

\section*{Biographies}
Xinyu Zhou is currently pursuing a Ph.D. degree at Nanyang Technological University (NTU) in Singapore. Her research interests include federated learning and Metaverse.

~ 

Jun Zhao is currently an Assistant Professor in the School of Computer Science and Engineering at Nanyang Technological University (NTU) in Singapore. He received a PhD degree in May 2015 in Electrical and Computer Engineering from Carnegie Mellon University (CMU) in the USA (advisors: Virgil Gligor, Osman Yagan; collaborator: Adrian Perrig), affiliating with CMU's CyLab Security \& Privacy Institute, and a bachelor's degree in July 2010 from Shanghai Jiao Tong University in China. Before joining NTU first as a postdoc with Xiaokui Xiao and then as a faculty member, he was a postdoc at Arizona State University as an Arizona Computing PostDoc Best Practices Fellow (advisors: Junshan Zhang, Vincent Poor). His research interests include federated learning, edge/fog computing, and Metaverse.

\vfill

\end{document}